%% file: acl2025.tex
\pdfoutput=1
\documentclass[11pt]{article}

\usepackage[final]{acl}

\usepackage{times}
\usepackage{latexsym}
\usepackage{amsmath} 
\usepackage[T1]{fontenc}
\usepackage[utf8]{inputenc}

\usepackage{microtype}

\usepackage{inconsolata}

\usepackage{graphicx}
\usepackage{multirow}
\usepackage{booktabs}
\usepackage{hyperref}
\usepackage{inconsolata}
\usepackage{pifont}% http://ctan.org/pkg/pifont

\usepackage{tabu}
\usepackage{fontawesome}
\usepackage{amsmath}
\usepackage{amsfonts}
\usepackage{amssymb}
\usepackage{enumitem}
\usepackage{listings}
\usepackage{xstring}
\usepackage{graphicx}
\usepackage{pbox}
\usepackage{subcaption}
\usepackage{epstopdf}
\usepackage{xstring}
\usepackage{multirow}

\usepackage{soul}
\usepackage{color}
\usepackage{graphicx}
\usepackage{multirow}
\usepackage{comment}
\usepackage{todonotes}
\usepackage{tcolorbox}
\usepackage{placeins}
\usepackage{listings} % Add this for using the listings package
\usepackage{graphicx}
\usepackage{subcaption}
\usepackage{verbatim}
\usepackage{stfloats}

\usepackage{xcolor} % For coloring text
\definecolor{lightblue}{rgb}{.50,.90,0.51}
\definecolor{tri}{rgb}{.25,.88,.82}
\definecolor{lilac}{rgb}{0.85,0.64,0.85}
\definecolor{atomictangerine}{rgb}{1.0, 0.6, 0.4}

\newcommand{\titu}{\emph{TituLLMs}}

\title{TituLLMs: A Family of Bangla LLMs with Comprehensive Benchmarking}

\author{
\textbf{Shahriar Kabir Nahin}\textsuperscript{1},
\textbf{Rabindra Nath Nandi}\textsuperscript{1},
\textbf{Sagor Sarker}\textsuperscript{1}, \\
\textbf{Quazi Sarwar Muhtaseem}\textsuperscript{1}, 
\textbf{Md Kowsher}\textsuperscript{2},
\textbf{Apu Chandraw Shill}\textsuperscript{1},
\textbf{Md Ibrahim}\textsuperscript{1}, \\
\textbf{Mehadi Hasan Menon}\textsuperscript{1},
\textbf{Tareq Al Muntasir}\textsuperscript{1}, 
\textbf{Firoj Alam}\textsuperscript{3}, \\
\textsuperscript{1}Hishab Singapore Pte. Ltd, Singapore, 
\textsuperscript{2}University of Central Florida, USA \\
\textsuperscript{3}Qatar Computing Research Institute, Qatar \\
\texttt{\{shahriar.nahin, rabindra.nandi, sagor.sarker\}@hishab.co, fialam@hbku.edu.qa}
}

\begin{document}
\maketitle
\begin{abstract}
In this paper, we present \titu{}, the \textit{first} large pretrained Bangla LLMs, available in 1b and 3b parameter sizes. Due to computational constraints during both training and inference, we focused on smaller models. To train \titu{}, we collected a pretraining dataset of approximately $\sim37$ billion tokens. We extended the Llama-3.2 tokenizer to incorporate language- and culture-specific knowledge, which also enables faster training and inference.
There was a lack of benchmarking datasets to benchmark LLMs for Bangla. To address this gap, we developed \textit{five benchmarking datasets}. We benchmarked various LLMs, including \titu{}, and demonstrated that \titu{} outperforms its initial multilingual versions. However, this is not always the case, highlighting the complexities of language adaptation. Our work lays the groundwork for adapting existing multilingual open models to other low-resource languages. To facilitate broader adoption and further research, we have made the \titu{} models and benchmarking datasets publicly available.\footnote{\href{https://github.com/hishab-nlp/titulm}{https://github.com/hishab-nlp/titulm}}

\end{abstract}

\input{sections/introduction}
 \input{sections/dataset}
\input{sections/experiments}
\input{sections/benchmarking}

\input{sections/results_and_discussion}

\input{sections/related_work}

\section{Conclusion}
\label{sec:conclusion}
% In this study, we present the \textit{first} pretrained Bangla LLMs (\textit{Titulm}) with $\sim37b$ tokens by adapting Llama-3.2 models. We have extended tokenizer to enrich language and culture specific knowledge, which also enables faster training and inference. Pretraining data collection is challenging for languages with low digital representation. We provide a complete recipe including raw web data collection, translation and synthetic data generation. Given that there was a lack of LLM based benchmakring dataset, therefore, we have developed five datasets consisting of $137k$ samples, covering knowledge and reasoning. The benchmakring dataset include both manual and our novel translation based approach. Using these dataset, we benchmarked different LLMs including \textit{Titulm}. We demonstrate that performs better in reasoning tasks without any instruction tuning. Future work include collecting large pretraining  datasets, and fine-tuning with instruction datasets. We have made our models publicly available for the community and for the reproduciblity, we aim to release training receipes and benchmarking datasets.
In this study, we present the \textit{first} pretrained Bangla LLMs, \titu{}, trained on $\sim37b$ tokens by adapting Llama-3.2 models. We extended the tokenizer to incorporate language- and culture-specific knowledge, which also enable faster training and inference. Pretraining data collection remains challenging for languages with low digital representation. To address this, we provide a comprehensive approach, including raw web data collection, translation, and synthetic data generation. Given the lack of LLM-based benchmarking datasets, we developed \textit{five datasets} comprising $137k$ samples, covering both \textit{knowledge} and \textit{reasoning}. The benchmarking dataset includes manually curated samples as well as a novel translation-based (EST) approach. Using these datasets, we benchmarked various LLMs, including \titu{}, demonstrating its superior performance in reasoning tasks without instruction tuning. Future work includes collecting larger pretraining datasets and fine-tuning with instruction-based datasets. 
% We have made our models publicly available, and to support reproducibility, we plan to release training recipes and benchmarking datasets.

% \textit{Titulm} series, 
% This paper presents significant advancements in developing monolingual Bangla language models through the \textit{Titulm} series.  
% By extending the tokenizer and leveraging a large-scale Bangla MMLU dataset in our benchmarking, we demonstrate the effectiveness of these models in enhancing word relation understanding and improving performance on commonsense and specialized reasoning tasks.  
% This work lays a strong foundation for the development of more capable models for Bangla and other low-resource languages, offering valuable resources for further research and application in multilingual LLMs. Additionally, we introduce a novel English to Bangla translation technique to enhance the translation data quality, contributing to more accurate and context-aware translations.

% Additionally, it highlights the potential for domain adaptation and the importance of creating high-quality instruction tuning datasets to further improve performance in specialized tasks.
 
\section{Limitations}
\label{sec:limitations}
There are two limitations in this work. Firstly, despite the improvements observed in the 3b variant, the model's performance on long contexts remains suboptimal, suggesting the need for further enhancement in handling extended sequences. Secondly, while the current models are trained solely on Bangla text, their performance could benefit from incorporating larger, English-centric datasets. This could facilitate better knowledge leveraging and potentially improve low-resource language performance, indicating a direction for future research. Since there is a lack of instruction tuning data in Bangla, we do not explore the full potential of instruction tuning, which could have further improved the models' performance on specialized tasks and domain adaptation.

\section*{Ethical Consideration}
\label{sec:ethics}
We do not anticipate any ethical concerns in this study. All datasets used were collected from publicly available sources, ensuring compliance with ethical research standards. No personally identifiable information (PII) was gathered or utilized in the development of our models. While we do not foresee any potential risks arising from the outcomes of this study, we strongly encourage users of the released models to adhere to responsible AI usage guidelines. This includes avoiding the generation or dissemination of harmful, misleading %, or biased 
content and ensuring that the models are employed in ethical and socially beneficial applications.

% \section*{Acknowledgments}

% Acknowledgments section here 

\bibliography{bib/bibliography}

\appendix

\input{sections/appendix}

\label{sec:appendix}

\end{document}

%% file: sections/introduction.tex
\section{Introduction}
\label{sec:introduction}

%https://arxiv.org/pdf/2406.14670
%https://arxiv.org/pdf/2404.05829

The rapid advancements in large language models (LLMs) have reshaped the field of AI, showcasing remarkable versatility across numerous tasks \cite{brown2020language,ouyang2022training,achiam2023gpt4,chowdhery2023palm}. These models demonstrate not only an ability to perform various NLP tasks but also an intriguing potential for self-assessment and continuous improvement \cite{liu2023gpteval,fu2023chain,chiang2023vicuna}. 
% As LLMs progress, their proficiency in self-reflection and correction has been the subject of considerable research \cite{bai2022constitutional,saunders2022self,madaan2023self,gou2024critic}.

Despite these advancements, LLM development—both open and closed—has predominantly focused on multilingual models, with a stronger emphasis on high-resource languages \cite{achiam2023gpt4,touvron2023llama}. While some models have extended coverage to medium- and low-resource languages \cite{le2023bloom,ustun2024aya,qwen2,gemma_2024}, their representation remains limited. Although some initiatives have aimed to train language-centric LLMs \cite{sengupta2023jais,fanar2024}, these efforts remain scarce due to the high costs associated with computational resources and data collection. Consequently, recent research has shifted towards adapting existing LLMs for new languages \cite{levine2024rakutenai,fanar2024}.

\begin{figure}[t]
  \centering
\includegraphics[width=0.85\columnwidth]{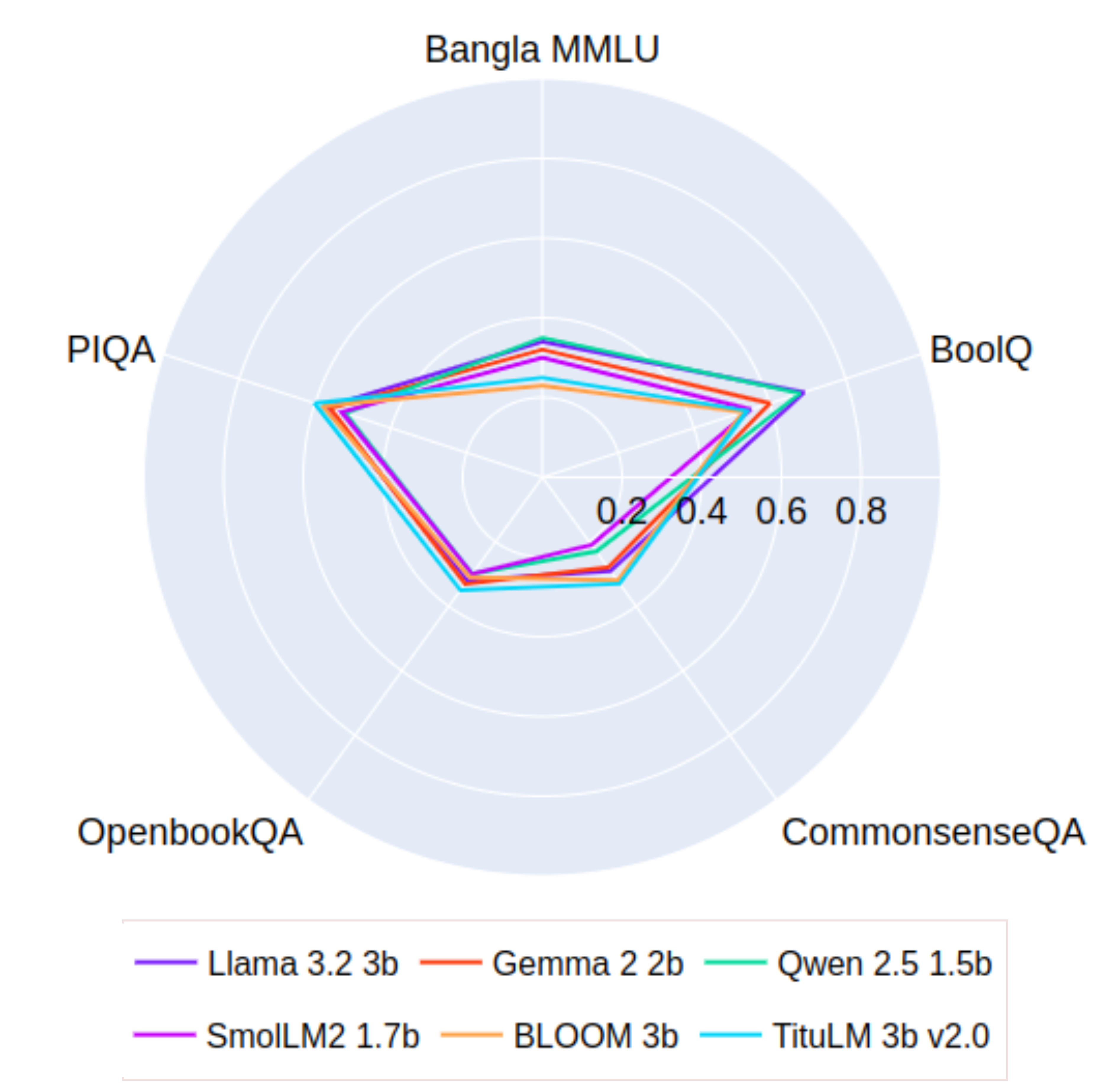}
  \caption{Performance scores per category TituLLM-3B and five other  models in 5-shot setting.}
  \vspace{-0.3cm}
  \label{fig:bangla_eval_data}
  \vspace{-0.3cm}
\end{figure}

Similarly, in the context of benchmarking LLMs, most efforts have primarily focused on high-resource languages~\cite{bang2023multitask, ahuja2023mega}, while low-resource languages, such as Bangla, have received limited attention~\cite{kabir2024benllm, zehady2024bongllama, bhattacharjee2023banglanlg}. \citet{zehady2024bongllama} developed LLMs for Bangla using Llama, leveraging only the Bangla subset of CulturaX~\cite{nguyen2024culturax}, which consists of 12.4 million diverse Bangla news articles. They further fine-tuned their model on 172k instruction samples from subsets of the Alpaca~\cite{taori2023stanford} and OpenOrca~\cite{lian2023openorca} datasets, which were translated from English. Their models were benchmarked on 120 queries across nine different generation tasks.

Addressing this gap is crucial, as linguistic and cultural diversity significantly impact language understanding and generation. Therefore, in this study, we focus on adapting existing LLMs (e.g., Llama) by expanding the model's vocabulary and performing continual pretraining. This process required extensive data collection from diverse sources. Given the relatively low availability of digital content in Bangla, we also developed synthesized datasets to supplement our training data.

Benchmarking LLM capabilities for Bangla remains challenging due to the lack of specialized datasets, particularly in areas such as world knowledge and commonsense reasoning. Although some efforts have been made to generate such datasets through translation \cite{lai-etal-2023-okapi}, they remain limited in scope. To address this gap, we have developed several native and translated datasets. \textbf{\textit{Compared to }}\citet{zehady2024bongllama}, our pretraining corpus is significantly larger ($\sim37b$ tokens) and is benchmarked on five different datasets covering world knowledge and commonsense reasoning, with a total dataset size of 132k entries.

% and evaluation have been largely focused on high-resource languages, 
% with comparatively less attention given to low-resource languages such as Bangla \cite{kabir2024benllm}. 

% To contribute to the growing body of LLM evaluation, we have developed a unique multiple-choice question (MCQ) and question-answer (QA) dataset specifically tailored for the low-resource Bangla language. This dataset is intended to benchmark various LLMs, evaluating their performance across essential tasks in Bangla and providing insights into their capacity for language comprehension, alignment, and safety in low-resource settings.

% \com{to do}
\begin{itemize}[noitemsep,topsep=0pt,labelsep=.5em] %leftmargin=*
\item We developed and released two models, \titu{}, adapted from Llama 3.2, which will enable future research. 
\item We provide a complete data collection recipe for pretraining LLMs including sources, approaches to synthetic data generation. 
\item We extended tokenizer to ingest language specific knowledge. 
% We prepare and release four types of datasets specifically designed for evaluating multilingual large language models (LLMs), including multiple-choice questions (MCQ) and question-answer (QA) pairs in Bangla;
\item We developed \textit{\textbf{five datasets}} to benchmark LLMs capabilities in terms of world knowledge, commonsense reasoning, and reading comprehension. Such datasets will serve as a first step to Benchmark LLMs for Bangla.
\item We proposed a novel translation techniques (\emph{expressive semantic translation}) that helps to develop high quality benchmarking dataset. 
\item Using the benchmaked datasets we benchmark various LLMs including \titu{} comparing performance across models to assess understanding of Bangla language. 
% to evaluate their capabilities in handling Bangla, a low-resource language, comparing performance across models to assess understanding of Bangla language structures and task-specific accuracy;
% \item We develop domain-generalization test sets that cover a wide range of linguistic domains, ensuring robust evaluation of how well LLMs generalize across different contexts and tasks in Bangla;
% \item We discuss the current limitations in evaluating LLMs for Bangla, including challenges with dialectal variations and the need for more comprehensive datasets. We also explore future research directions to improve model performance and expand dataset diversity for Bangla. 
\end{itemize}

\noindent
Our study reveals several interesting findings:

\noindent
\textbf{Vocabulary Extension:} We explore the impact of vocabulary extensions on the base Llama Tokenizer by increasing the number of new tokens from 32K to 96K in increments of 16K. We found that average tokens per word (TPW) decreases as the number of tokens increases up to a certain point, after which it declines only minimally. Therefore, when adding new tokens, we must also consider the fertility rate to balance the trade-off between training and inference.

% \noindent
% \textbf{Performance of Adaptive Models:} Our analysis reveals that existing models exhibit stronger predictive capabilities when evaluated on code-mixed and knowledge-intensive data than original Bangla data. However, \titu{} outperforms other models on reasoning tasks in pure Bangla datasets as illustrated in Figure \ref{fig:bangla_eval_data}. This superior performance can be attributed to its tokenizer and pretraining corpus, which includes a substantial volume of cleaned Bangla text, enabling more effective language understanding.

\noindent
\textbf{Commonsense Capability:} \titu{} demonstrates strong commonsense knowledge but has limited capability in world knowledge (e.g., Bangla MMLU). Further training with instruction fine-tuning may enhance its performance in this area.

%% file: sections/dataset.tex
\section{Pretraining Data}
\label{sec:datasets}
\begin{figure*}[h]
    \centering
    \includegraphics[width=1.0\linewidth]{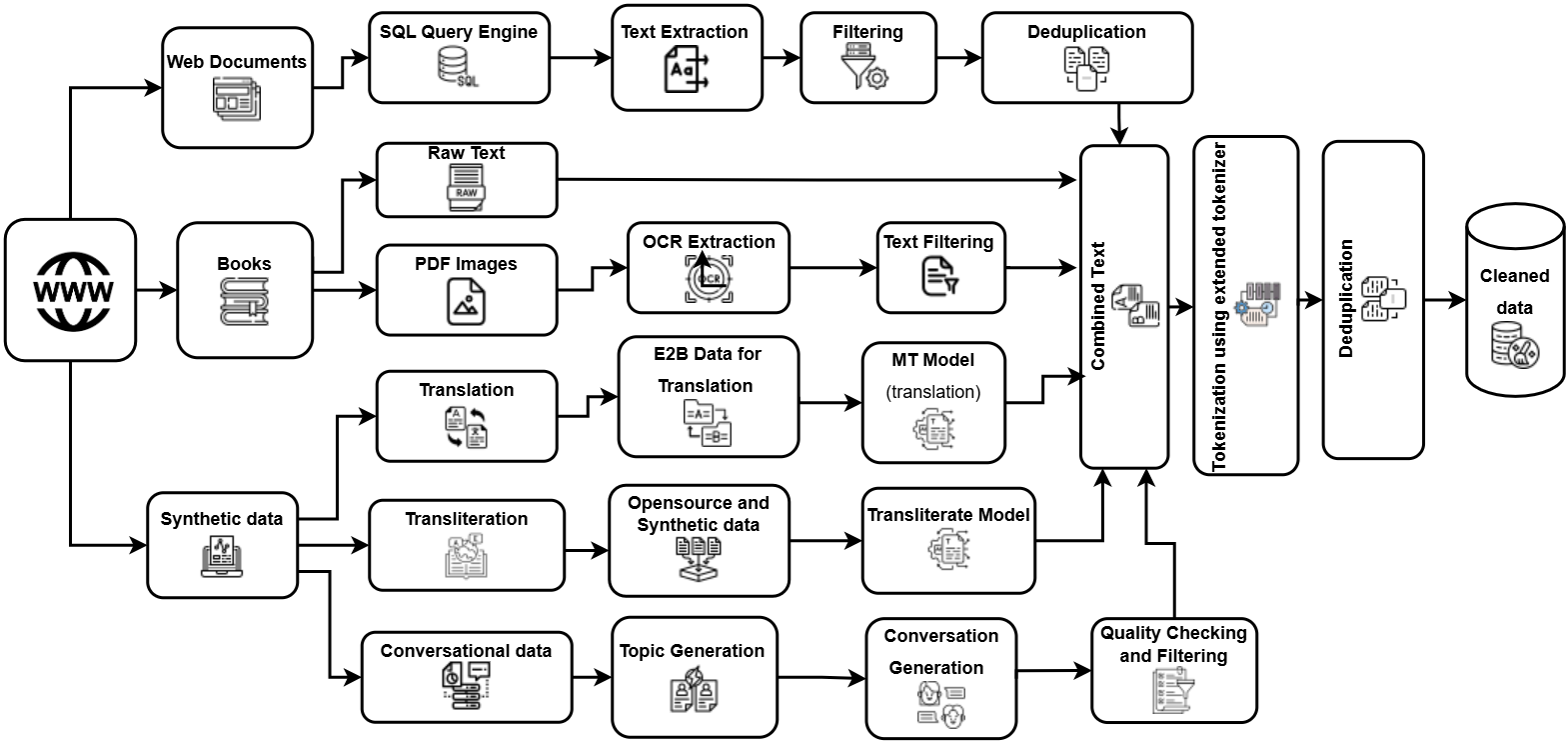}
    \vspace{-0.2cm}
    \caption{Overview of the pretraining data collection and preprocessing pipeline -- A workflow illustrating the steps involved in gathering, filtering, and preparing data for LLM pretraining.
    }
    \label{fig:data_collection_final}
    \vspace{-0.3cm}
\end{figure*}

Pretraining data for Bangla is very limited compared to very high quality data available for English and other high resource languages \cite{penedo2023refinedweb,soldaini2024dolma}. Hence, we needed collect pretraining dataset for training \titu{}. We have compiled a substantial Bangla raw dataset from a diverse range of sources, encompassing both formal and informal linguistic styles. The dataset is primarily derived from three key sources: web documents, books, and synthetically generated text. The synthetically generated data includes translated data, transliterated data, audio transcripted data, etc. An outline of our data collection and preprocessing pipeline is shown in Figure  \ref{fig:data_collection_final}. The final high-quality dataset amounts to approximately 268 GB, with 22 GB designated as a validation set, sampled proportionally to maintain the heterogeneity of the data. In total, the corpus contains $\sim37$ billion tokens, optimized for training and evaluation across various Bangla language processing applications. In Table~\ref{tab:titulm_tokens} (in Appendix) and in Figure \ref{fig:pretraining_data_distribution}, we report the distribution of tokens for different sources.

% , we  briefly describes our pretraining data collection pipeline. 
%64.19 

% \begin{table}[h]
%     \centering
%     \begin{tabular}{l c}
%         \hline
%         \textbf{Data} & \textbf{GPT-4o tokens (B)} \\
%         \hline
%         Web Documents & 46.84 \\
%         Books & 9.06 \\
%         Synthetic Data & \\
%         \quad Translated & 2.14 \\
%         \quad Romanized & 3.87 \\
%         \quad Conversation & 0.54 \\
%         \quad Audio Transcription & 1.74 \\
%         \hline
%         \textbf{Total} & \textbf{64.19} \\
%         \hline
%     \end{tabular}
%     \caption{GPT Tokens Count for Different Data Sources}
%     \label{tab:gpt_tokens}
% \end{table}

% \begin{figure}[h]
%     \centering
%     \includegraphics[width=\linewidth]{figures/pretraining-data-collection-pipeline.png}
%     \caption{Pipeline for pertaining data collection.}
%     \label{fig:Pipeline for pertaining data collection}
% \end{figure}

\begin{figure}[t]
  \centering
  \includegraphics[width=0.7\columnwidth]{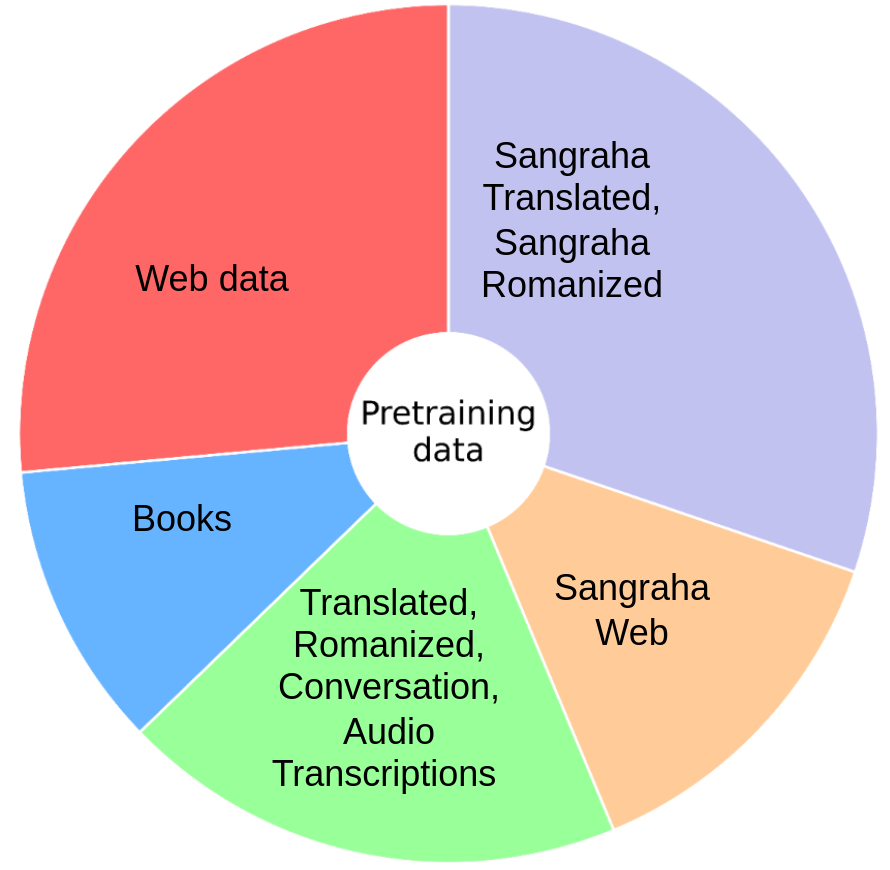}
  \vspace{-0.3cm}
  \caption{The pretraining dataset consists of $\sim37b$ billion tokens distributed across various sources: Web (9.8b), Books (4b), Synthetic Data (7.06b), Sangraha Web (5b), and Sangraha Synthetic (10.94b). Synthetic Data includes Translated (1.47b), Romanized (3.87b), Conversation (0.42b), and Audio Transcription (1.30b), while Sangraha Synthetic comprises Translated (4.26b) and Romanized (6.68b).}
  \label{fig:pretraining_data_distribution}
  \vspace{-0.4cm}
\end{figure}

\subsection{Web Documents}
We curated the \textit{Common Crawl}~\cite{raffel2020exploring} dataset and followed multiple steps to extract and clean the final dataset, as illustrated in Figure \ref{fig:data_collection_final}. Below, we briefly discuss each step. 

\noindent \textbf{SQL Query Engine:} Using Amazon Athena,\footnote{\url{https://aws.amazon.com/athena/}} we queried the vast Common Crawl dataset to isolate Bangla-specific HTML data and URLs. We applied filtering based on content language, URL patterns indicative of Bangla domains (e.g., \texttt{.gov.bd}), and host information, covering data from 2017 to 2024.  

\noindent \textbf{Text Extraction:} We used \textit{Trafilatura}~\cite{barbaresi-2021-trafilatura} tool for its effectiveness in extracting structured, clean text from HTML. This step preserved the structure of text patterns while minimizing extraneous noise, ensuring the retention of original content nuances.

\noindent \textbf{Filtering:} To assess the usability and relevance of the extracted text, we generated 18 distinct hand-crafted rules specifically designed for the Bangla language. These rules evaluated various aspects of text quality, including sentence completeness, document structure, and content appropriateness, leveraging Bangla-specific tokenizers for a proper filtering process. The details of these rules are discussed in Section \ref{ssec:rule_data_filtering}. After extraction, documents were assessed against predefined thresholds associated with each rule. Based on the threshold we filtered documents that did not pass the criteria. 
% %and tools for precise language handling.  
% using a \com{Gopher Rule} approach to ensure high standards. Based on these evaluations, documents were categorized into 'pass' and 'fail' groups, effectively segregating high-quality content for further analysis and applications.  

\noindent \textbf{Deduplication:} We applied the MinHash deduplication algorithm~\cite{kocetkov2022stack} to the filtered web dataset to eliminate redundant content.  

% This methodical approach not only streamlined the extraction of targeted content but also ensured the refinement of Bangla text, making it suitable for varied linguistic applications and analyses.

\subsection{Books}
We have compiled a diverse collection of open-source Bangla books, primarily in PDF format, spanning a broad temporal range from historical to contemporary works. Below we briefly discuss the steps taken to extract the text from the book collection.

\noindent \paragraph{Raw Text:} For digitally available texts, we directly extract the machine-readable content, requiring minimal processing due to the already high quality of these formats.

\noindent \paragraph{PDF Images:} 
To digitize a vast collection of non-digital and older texts from books, we utilize two leading Optical Character Recognition (OCR) systems: Google OCR\footnote{\url{https://cloud.google.com/use-cases/ocr}} and Tesseract\footnote{\url{https://github.com/tesseract-ocr/tesseract}}. We used Tesseract to reduce the cost. These texts, often derived from sources that have deteriorated over time or originated in non-digital formats, pose significant challenges in terms of quality and legibility. To extract high-quality text, we implement carefully designed techniques comprising several steps.

\noindent \textbf{Text extraction using Google OCR:} For the majority of books, we utilized Google OCR to extract text. Although the OCR accuracy was generally high for more recent books, the quality varied significantly for older books. Identifying and filtering out poor-quality text from a large and diverse collection proved to be a challenging task. To address this issue, we applied several quality-control techniques: \textit{(i)} using KenLM \cite{heafield2011kenlm} to filter noisy text based on ranking, \textit{(ii)} evaluating the average number of words and sentences per page, and \textit{(iii)} calculating the percentage of correct Bangla words. Details are discussed in Section \ref{ssec:app_rules_google_ocr}.

\noindent \textbf{Document Segmentation and Tesseract OCR:} We performed document segmentation alongside OCR for a smaller portion of the collected books. Initially, we trained a YOLO segmentation model on a Bangla document segmentation dataset \cite{shihab2023badladlargemultidomainbengali} to identify and classify different components of the documents (e.g., text boxes, tables, paragraphs, and images). We removed complex sections such as tables and images, as they were not relevant to the text extraction process. Subsequently, we applied Tesseract OCR to the remaining document text and repeated the filtering process outlined earlier. In addition to the filtering steps, we introduced an additional measure: we calculated the number of words with more than 80\% confidence and set a threshold at the 95th percentile to filter out low-quality text. After applying all these processes, 50\% of the initial data was retained.

\subsection{Synthetic Data}
Due to the low representation of digital content in Bangla, we have developed a large-scale synthetic dataset for Bangla, which include transcription, translation and transliterated data. 

% \subsubsection{Transcribed Text}
\paragraph{Transcribed Text:}  
We collected conversational and spoken language data transcribed using the Bangla Automatic Speech Recognition (ASR) system~\cite{nandi-etal-2023-pseudo}. This system enables us to capture various colloquial and regional linguistic variations in Bangla. We collected approximately 56k hours of speech data from diverse online sources. All collected speech data were transcribed using the ASR system.

\paragraph{Translation Data:}
% \com{it is not clear why you also used The Llama-3.1-8B-Instruct for translation? Was not the trained MT model enough?} 

% \com{From Nahin: It wasn't enough because most of the open-source translated data was synthetically translated from English to Bengali. We wanted to ensure better quality and quantity. That's why we have curated clean Bengali text and translated Bengali text to English text. As the Bengali text is already curated from Wikipedia and news sources, they are good for training. Additionally, we found that llama3.1-8b can generate English text well rather than Bengali. As a result, we get a high-quality Bengali-English translation pair for training.}

To collect English-to-Bangla translated data, we trained an NLLB-based (600M-Distilled) model \cite{nllbteam2022languageleftbehindscaling} with the goal of developing a smaller, language-pair-specific (en-bn) model. We decided to train a translation model because our observations indicate that currently available multilingual models, such as Llama-3.1-8B-Instruct, have limited capability for Bangla-specific generation tasks. However, they have shown superior performance in English-specific generation.
% Additionally, our observations indicate that currently available multilingual models, such as Llama-3.1-8B-Instruct, are not sufficiently accurate for Bangla-specific generation tasks.

For training the en-bn machine translation (MT) model, we collected open-source translation data from various platforms, including BanglaNMT \cite{hasan-etal-2020-low} and Samanantar \cite{ramesh-etal-2022-samanantar}. Furthermore, we generated synthetic bn-en translation pairs using Bangla news sources and Wikipedia as source data, employing Llama-3.1-8B-Instruct ~\cite{touvron2023llama} for target data generation. We selected Llama for this task due to its superior English-language capabilities.
Using this approach, we created a dataset comprising approximately 60 million translation pairs, which we then used to train the NLLB model. We have named this model as \textit{Titu-Translator}. On our in-house test dataset, the BLEU score is 37.6. %of this model is 37.6. 

% Other than BLEU score, 
\noindent
\textbf{Evaluation using Scalar Quality Metrics (SQM).} We have also evaluated our model using SQM \cite{lankford2022human}. It is a structured and interpretable human evaluation of machine translation.  Table \ref{tab:pretrain-trans-human-eval} presents the results of human evaluation conducted by two evaluators across five distinct domains: Business, Entertainment, Politics, Tech, and Sport. Among the evaluated models, \textit{Titu-Translator} demonstrates strong performance, achieving an overall average score of 4.99, which places it among the top-performing models. It performs particularly well in the Tech and Entertainment domains, with scores up to 5.30 from Human-2 and 4.95 from Human-1. While Indic-Trans-2 slightly outperformed \textit{Titu-Translator}, we chose \textit{Titu-Translator} for synthetic pretraining due to its lightweight design and alignment with Bangladeshi linguistic styles.
% It shows especially strong results in the Tech and Entertainment domains, with scores reaching as high as 5.30 from Human-2 and 4.95 from Human-1. However, Indic-Trans-2 performed slightly better than \textit{Titu-Translator}. But because of the lightweightness and Bangladeshi linguistic styles of \textit{Titu-Translator}, we have used \textit{Titu-Translator} to generate synthetic pretraining data. 
Appendix \ref{sec:translation_model_eval} describes more about the evaluation process.

Once the model has been trained, we have used it to translate a corpus of English news articles\footnote{\href{https://www.kaggle.com/datasets/davidmckinley/all-the-news-dataset}{https://www.kaggle.com/datasets/davidmckinley/all-the-news-dataset}} into Bangla.

Additionally, we added translated data from the \emph{Sangraha} dataset~\cite{khan2024indicllmsuite} to our corpus. The synthetic data coming from the \emph{Sangraha} dataset is generated with Indic-Trans-2, which is reported to be the top-performing model for English to Bangla translation in their report \cite{gala2023indictrans2highqualityaccessiblemachine}. It is also reflected in our SQM evaluation.

\begin{table*}[htb!]
\centering
\setlength{\tabcolsep}{2pt} 
\scalebox{0.83}{%
\begin{tabular}{lcc|cc|cc|cc|cc|c}
\toprule
\textbf{Model} &
\multicolumn{2}{c|}{\textbf{Business}} &
\multicolumn{2}{c|}{\textbf{Entertainment}} &
\multicolumn{2}{c|}{\textbf{Politics}} &
\multicolumn{2}{c|}{\textbf{Tech}} &
\multicolumn{2}{c|}{\textbf{Sport}} &
\textbf{Average} \\
 & H1 & H2 & H1 & H2 & H1 & H2 & H1 & H2 & H1 & H2 & \\
\midrule
Titu Translator     & 4.8 & 5.3 & 4.85 & \textbf{5.4} & 4.8 & 5.05 & 4.95 & \textbf{5.3} & 4.75 & 4.65 & 4.99 \\
nllb-600M           & 4.45 & 5.1 & 4.65 & 5.0 & 4.75 & 4.85 & 4.85 & 5.05 & 4.55 & 4.2 & 4.75 \\
csebuetnlp-t5       & 4.55 & 4.95 & 4.65 & 4.8 & \textbf{4.85} & 4.9 & 4.6 & 4.9 & 4.65 & 4.65 & 4.75 \\
Indic-Trans-2       & \textbf{4.95} & \textbf{5.35} & 4.85 & \textbf{5.35} & \textbf{4.85} & \textbf{5.35} & \textbf{5.0} & 4.9 & \textbf{5.0} & \textbf{5.05} & \textbf{5.07} \\
Google Translator   & 4.6 & 4.75 & \textbf{4.89} & 4.65 & 4.65 & 4.85 & 4.95 & 4.8 & \textbf{5.0} & 4.55 & 4.77 \\
\bottomrule
\end{tabular}
}
\vspace{-0.2cm}
\caption{Human (H) evaluation by two annotators across five domains for each translation model.}
\label{tab:pretrain-trans-human-eval}
\vspace{-0.2cm}
\end{table*}

% we have collected and trained our model with 

\paragraph{Transliteration Data:}
The use of romanized text is very common in everyday communication for Bangla~\cite{fahim2024banglatlit}. To address this, we have developed a Bangla-to-Romanized dataset by training an NLLB-based (600M-Distilled) model. For model training, we collected transliteration pairs from the \textit{Sangraha} dataset and generated additional synthetic transliteration pairs using the GPT-4 model~\cite{achiam2023gpt4}. We then used this dataset to train the NLLB-based transliteration model. The BLEU score for this model is 65.1, as evaluated on an in-house test dataset. We then used this model to create the transliteration dataset by selecting a small subset of collected Bangla Wikipedia articles.

% We have also prepared a Bengali-to-Romanized Bengali transliteration dataset for training the NLLB-600M-Distilled model. \com{not clear -- The dataset includes both synthetic transliteration pairs sourced from the Sangraha Dataset and additional synthetic transliteration pairs generated using the GPT-4 model~\cite{achiam2023gpt4}.}

% \com{From Nahin: Synthetic data from Sangraha mostly contains Indian transliteration style, whereas we have generated some transliteration pair that follows Bangladeshi transliteration style. For example, 'যুব' is spelled as 'Yuva' in Indian style but in Bangladesh it is transliterated as 'Jubo'.}

% \cite{Nandi: vai, taking information from the team} 

% and used the trained model to produce a transliterated text corpus. This corpus was then incorporated into the final dataset for further processing and model training.

% \cite{openai2024gpt4technicalreport}. 

\paragraph{Conversational Data:}
To enhance the model with conversational capabilities, we enriched our dataset by incorporating conversational data. We have crawled topics (e.g.,  \textit{``Rabindranath Tagore's contributions
to Bengali art''}) from Wikipedia and Banglapedia on which we generated conversations between two agents. To achieve this we have developed an agentic system where two agents interact with each other on a given topic. In Appendix Table \ref{tab:agent_details}, we have provided examples of topic, roles, and a detail of the prompt. An example conversation is provided in Appendix Figure \ref{fig:data_collection_final}. The average number of turns per conversation is 8. In total, we added $\sim$1 million conversations to the dataset. 
% The system was an automated system where agents were engaged in discussion with each other on various topics. 
% To maintain the quality of the dataset we have used a complex agentic system here which can work as a self-evaluator of the generated conversations. 

\subsection{Sangraha Dataset}
Additionally, we enriched our dataset by integrating the open-source Sangraha dataset. It is the largest high-quality, cleaned Indic language corpus. We incorporated the Bangla web data portion of the dataset into our training set.

%% file: sections/experiments.tex
\section{Pretraining}
\label{sec:pretraining}

\subsection{Tokenizer Training}
We developed a custom tokenizer for Bangla text using Tiktoken\footnote{\href{https://github.com/openai/tiktoken}{https://github.com/openai/tiktoken}}, which employs Byte Pair Encoding (BPE)~\cite{sennrich2015neural} for tokenization. To train this tokenizer, we sampled 48 GB of data from our pretraining Bangla corpus. Additionally, we modified the original Tiktoken codebase to enhance its efficiency and better accommodate the morphological complexities of Bangla. After training multiple tokenizers on the same subset, we merged each newly trained tokenizer with the existing Llama-3.2 tokenizer. This merging process aimed to preserve the strengths of the original tokenizer while integrating domain-specific vocabulary and improving segmentation for Bangla. To evaluate the performance of each merged tokenizer, we computed the average tokens per word (TPW) on a separate 1 GB sample from the original corpus. Table~\ref{tab:tokens-per-word}, in Appendix, summarizes the TPW values for both the original Llama-3.2 tokenizer and the newly trained tokenizers. A lower TPW generally indicates more efficient segmentation, which reduces sequence lengths and may enhance downstream model performance.

We trained five tokenizers with different vocabulary sizes, as presented in Table~\ref{tab:tokens-per-word}. Each of these tokenizers was then merged with the Llama-3.2 tokenizer to create five new tokenizers. Notably, the Llama-3.2 tokenizer exhibits a very high TPW value, which affects its performance for Bangla. In contrast, the newly developed tokenizers demonstrate significantly lower TPW values.

The table also shows that increasing the vocabulary size of the new tokenizers generally results in a lower TPW count. However, the relationship between vocabulary size and TPW is not strictly linear. While TPW decreases with larger vocabularies, the reduction becomes less significant for tokenizers with very large vocabulary sizes.

% We have trained five tokenizers with five different vocabulary sizes as presented in Table~\ref{tab:tokens-per-word}, which were then merged with llama-3.2 tokenizer to prepare five new tokenizers. We can notice that the llama-3.2 tokenizer has a very high TPW value. As a result, the performance of llama-3.2 is not up to the mark for Bangla. Additionally, newly prepared tokenizers have a very low TPW. The table also indicates that if we increase the vocabulary size of new tokenizers, then the tokenizer's TPW count decreases. However, the token per word doesn't maintain a linear relation with vocabulary. Rather, the reduction of TPW is not significant for tokenizers with large vocabularies. 

% They are 32K, 48K, 64K, 80K, and 96K. 
% Then these tokenizers are merged with llama-3.2 tokenizer to prepare five new tokenizers: llama-3.2-plus-32K, llama-3.2-plus-48K, llama-3.2-plus-64K, llama-3.2-plus-80K, and llama-3.2-plus-96K.  

\subsection{Model Architecture}
We have modified Llama-3.2-1b and Llama-3.2-3b models according to the merged tokenizers. As too many new tokens will increase the model's size and the training complexity, we modified the models according to Llama-3.2-plus-48K tokenizer. We added extra embedding vectors in the embedding layer and modified the lm-head according to the vocabulary size.

\subsection{Pretraining}
After modifying the models, we pre-trained them on our full dataset using LlamaFactory~\cite{zheng2024llamafactory}. Both models were trained with a context length of 4096, with packing enabled for maximum efficiency. Training for one epoch required 1,750 H100 GPU hours. 
% Table~\ref{tab:titulm_tokens} presents the TituLM token counts from different sources.
% training repository for this process. 
% The training was conducted on our full Bangla corpus, which contains approximately 36.8 billion Llama-3.2-Plus-48K tokens. 

% \begin{table}[h]
%     \centering
%     \begin{tabular}{p{4cm} c}
%         \hline
%         \textbf{Data} & \textbf{TituLM tokens (B)} \\
%         \hline
%         Common Crawl, Books, and Sangraha Deduplicated Data & 18.80 \\ \midrule
%         Our Synthetic Data & \\
%         \quad Translated & 1.47 \\
%         \quad Romanized & 3.87 \\
%         \quad Conversation & 0.42 \\
%         \quad Audio Transcription & 1.30 \\ \midrule
%         Sangraha & \\
%         \quad Translated & 4.26 \\
%         \quad Romanized & 6.68 \\
%         \hline
%         \textbf{Total} & \textbf{36.80} \\
%         \hline
%     \end{tabular}
%     \caption{TituLM Tokens Count for Pretraining.}
%     \label{tab:titulm_tokens}
% \end{table}

% \label{sec:experiments}

% \subsection{Data Splits}
% #To DO Nahin 
% \subsection{Training data}

% \subsection{Evaluation data}
% \subsection{Models}
% #TODO Nahin 
% \subsection{Continual Pretraining}
% #TODO Nahin 

% \subsection{Evalution}
% Evaluation Hareness Framework 
% #sagor 

%% file: sections/benchmarking.tex
\section{Evaluation}

\subsection{Evaluation Setup}
For evaluation, we utilized the lm-evaluation-harness.\footnote{\url{https://github.com/EleutherAI/lm-evaluation-harness}} We used normalized accuracy as a metric. Our assessment focuses on key aspects such as \textit{knowledge} and \textit{reasoning}.

%%%%%%%%%%%%%%%%%%%%%%%%%%%%%%%

%%%%%%%%%%%%%%%%%%%%%%%%%%%%%%%%%

\subsection{Banchmarking Datasets}
We benchmarked \titu{} alongside other popular LLMs using five newly prepared evaluation datasets. The dataset is composed of multiple subsets of the benchmarking set, including Bangla MMLU (87,869 entries), Piqa BN (17,177 entries), CommonsenseQA BN (10,962 entries), OpenBookQA BN (5,944 entries), and BoolQ BN (1,976 entries). This distribution highlights the significant dominance of the Bangla MMLU subset within the overall evaluation dataset.

Below, we describe the development process for each dataset. Table~\ref{tab:benchmarking_dataset} presents the distribution and splits of each dataset.

\begin{figure}[htb!]
\centering
\includegraphics[width=0.65\columnwidth]{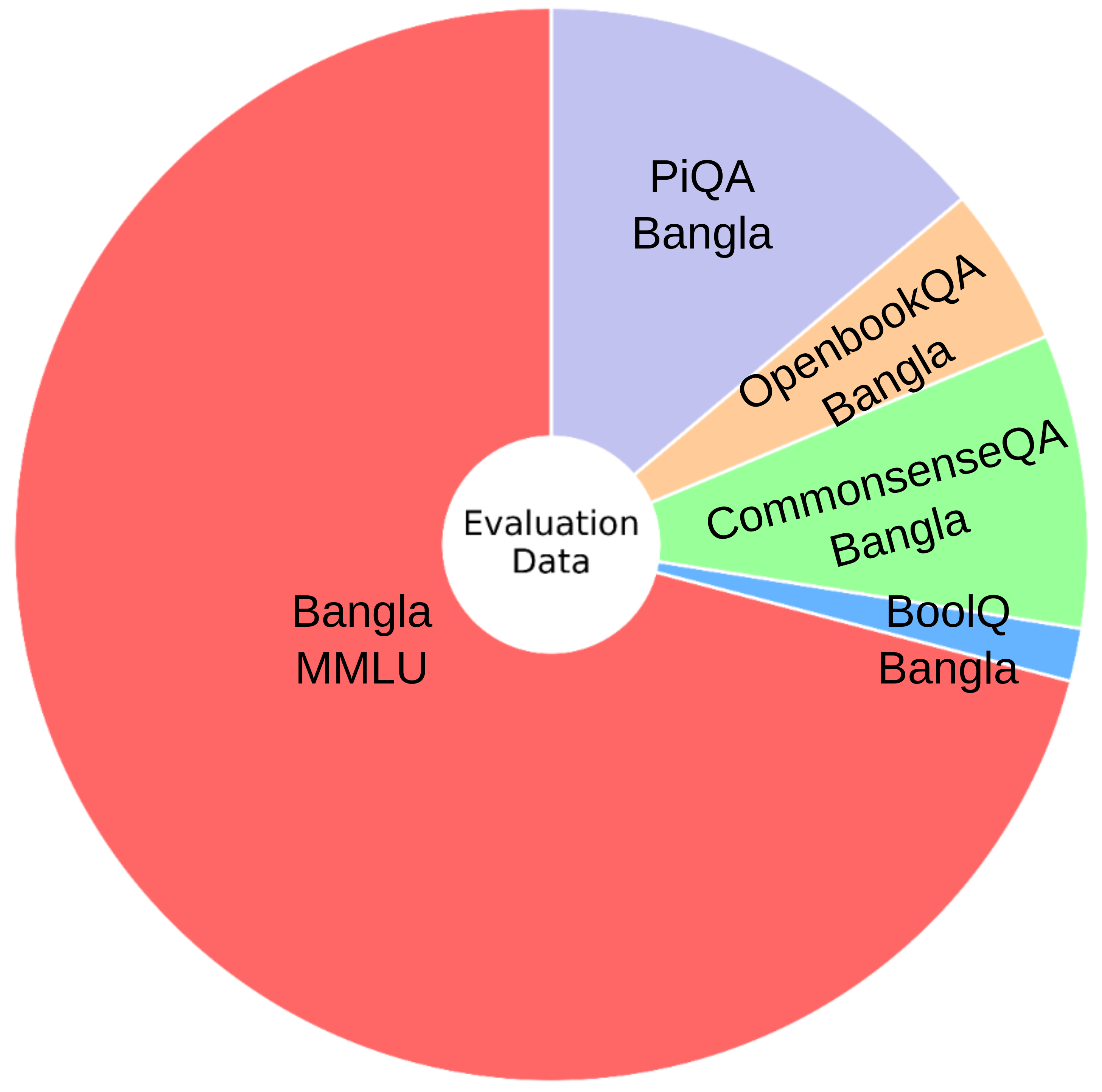}
\vspace{-0.3cm}
\caption{Distribution of an benchmarking dataset totaling $\sim124$ entries.
% distribution of an evaluation dataset totaling 131,928 entries across various subsets: Bangla MMLU with 87,869 entries, Piqa BN with 17,177 entries, Commonsenseqa BN with 10,962 entries, Openbookqa BN with 5,944 entries, and Boolq BN with 1,976 entries. 
% This highlights the significant representation of the Bangla MMLU subset within the evaluation data.
}
\label{fig:bangla_eval_data}
\vspace{-0.3cm}
\end{figure}

% We evaluated our pre-trained models on both Bangla and English benchmark datasets. Despite being primarily trained on Bangla data, the model’s English capabilities were assessed to explore its cross-lingual potential. 
% In this section, we report the results for \textbf{titulm-gemma-2-2b-v1.0} models on standard automatic benchmarks. All evaluations were conducted using the \texttt{lm-evaluation-harness} library to ensure consistency and reproducibility.

% \begin{itemize}[noitemsep,topsep=0pt,labelsep=.5em]
\noindent \textbf{Bangla MMLU:} 
% A proprietary multiple-choice question dataset developed by Hishab, curated 
We curated multiple-choice questions from various open-source educational websites
% , such as Aapathshala\footnote{\url{https://aapathshala.com/}} 
and textbooks, inspired by the original MMLU dataset~\cite{hendrycks2020measuring}. The dataset includes multiple-choice questions from different Bangladeshi exams, such as job exams, the Bangladesh Civil Service Exam, and undergraduate admission exams.

\begin{figure}[htb!]
  \centering
  \includegraphics[width=0.8\columnwidth]{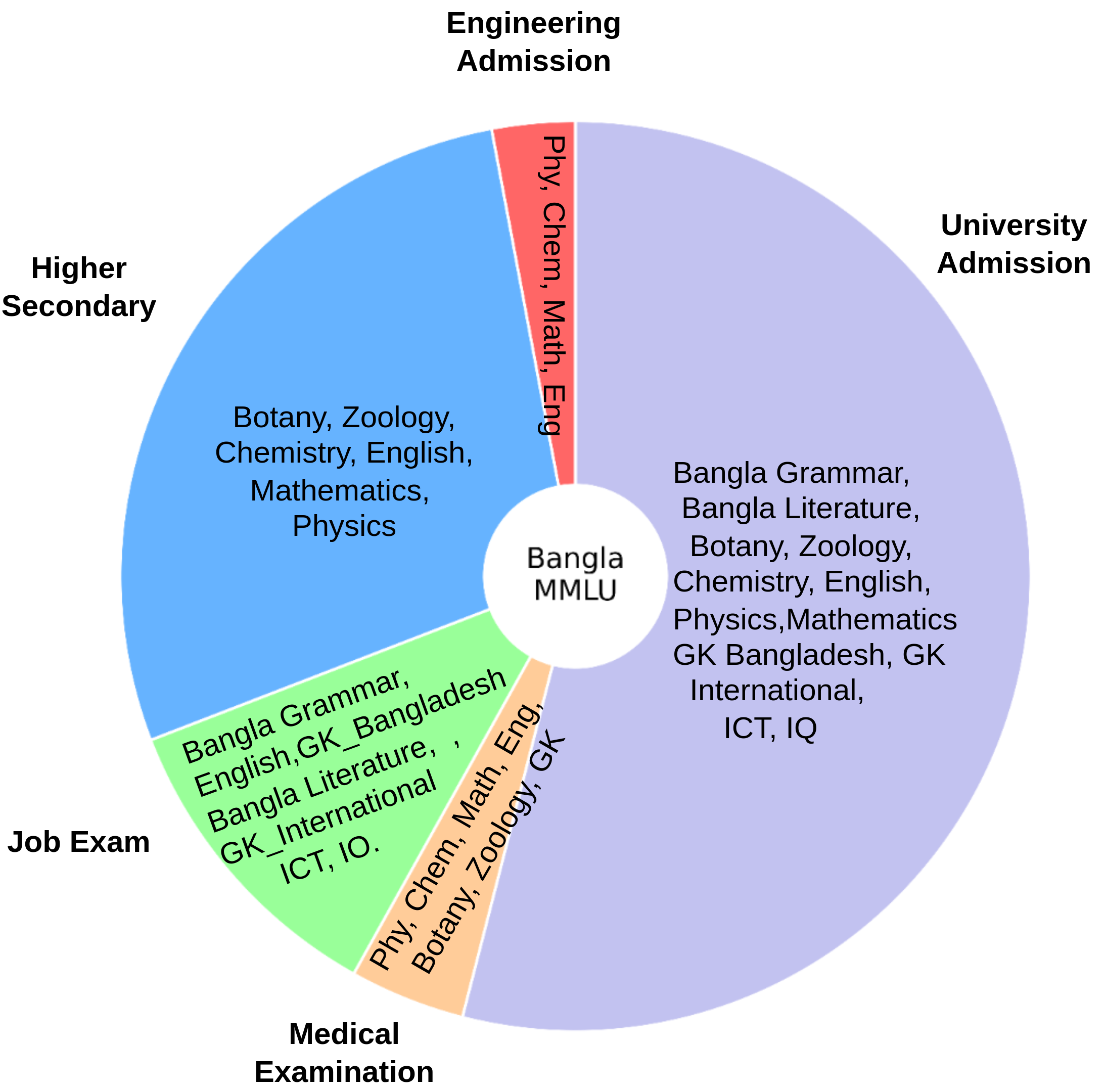}
  \caption{The Bangla MMLU dataset comprises a total of 87,869 questions distributed across various educational categories: University Admission (47,394), Higher Secondary (25,437), Job Exams (9,122), Medical Admission (3,764), and Engineering Admission (2,152).}
  \label{fig:bangla_mmlu_data}
  \vspace{-0.3cm}
\end{figure}

Figure \ref{fig:bangla_mmlu_data} provides a detailed breakdown of the Bangla MMLU dataset, which contains 87,869 questions spanning multiple educational categories. These include University Admission (47,394), Higher Secondary (25,437), Job Exams (9,122), Medical Admission (3,764), and Engineering Admission (2,152). The dataset reflects a diverse range of question types relevant to various levels of academic and professional assessments, making it a comprehensive benchmark for evaluating LLMs in Bangla educational contexts.

\noindent  \textbf{CommonsenseQA Bangla (CSQA):} We translated the CommonsenseQA dataset~\cite{talmor2018commonsenseqa} into Bangla using our custom translation-based approach, \textit{Expressive Semantic Translation (EST)}. This method generates multiple translations for a sentence and iteratively refines them to select the most suitable version. More details on this approach are discussed in Appendix Section~\ref{sec:app_est}.

% A Bangla translation of the CommonsenseQA dataset \cite{talmor2018commonsenseqa}, utilizing a novel \textit{Expressive Semantic Translation (EST)}(described in appendix) method. This method takes multiple translations for a sentence and iteratively refines to get most candidate version. 

\noindent \textbf{OpenBookQA Bangla (OBQA):} This dataset is a Bangla translation of the OpenBookQA dataset~\cite{mihaylov2018can}, which is designed to test a model's ability to apply elementary science knowledge to answer open-domain multiple-choice questions. We translated OpenBookQA into Bangla using our EST method.

\noindent  \textbf{PIQA Bangla (PIQA):} This dataset is a Bangla translation of the Physical Interaction: Question Answering (PIQA) dataset~\cite{bisk2020piqa}, which evaluates a model's understanding of everyday physical reasoning and common-sense interactions. PIQA consists of multiple-choice questions requiring knowledge about how objects interact in the real world, such as choosing the most practical way to perform a given task. For the translation, we used our EST method. 

\noindent  \textbf{BoolQ Bangla (BoolQ):} This dataset is inspired by BoolQ~\cite{clark2019boolq}, a reading comprehension benchmark that evaluates a model's ability to answer yes/no questions based on a given passage. The dataset consists of triplets in the form of (question, passage, answer). Passages were sourced from Bangla Wikipedia, Banglapedia, and news articles, ensuring a diverse range of topics and contexts. To generate high-quality questions and answers, we leveraged GPT-4.

\begin{table}[h]
\centering
\setlength{\tabcolsep}{3pt} 
\scalebox{0.75}{%
\begin{tabular}{@{}llrrrr@{}}
\toprule
\textbf{Dataset} & \textbf{Method} & \textbf{Train} & \textbf{Val.} & \textbf{Test} & \textbf{Dev} \\ 
\midrule
Bangla MMLU & Manual & - & 72,944 & \textbf{14,750} & 175 \\ 
BoolQ & GPT-4 & 815 & 432 & \textbf{729} & - \\ 
CommonsenseQA & EST & 9,741 & \textbf{1,221} & - & - \\
OpenBookQA & EST & 4,947 & 500 & \textbf{497} & - \\
PIQA & EST & 15,339 & \textbf{1,838} & - & - \\ 
\bottomrule
\end{tabular}
}
\vspace{-0.2cm}
\caption{Data splits and distribution of the Benchmark dataset. Val.: Validation.}
\label{tab:benchmarking_dataset}
\vspace{-0.3cm}
\end{table}

%% file: sections/results_and_discussion.tex
\section{Results and Discussion}
\label{sec:results}

We evaluate each model in 0-shot and 5-shot settings to assess their few-shot adaptability. Table~\ref{tab:benchmark-results} presents the detailed results for all models,\footnote{GPT~\cite{openai2023gpt4}, Llama~\cite{touvron2023llama}, Gemma~\cite{team2024gemma}, Qwen~\cite{chu2024qwen2}, SmolLM2~\cite{allal2025smollm2}, BLOOM~\cite{le2023bloom}, BongLLaMA~\cite{zehady2024bongllama}.} including the \titu{} variants. Table \ref{tab:benchmark-results} shows the accuracy of various models with less than or equal to 3b parameters and the GPT-davinci-002 model.

\begin{table}[htb!]
\centering
\setlength{\tabcolsep}{2pt} 
\scalebox{0.73}{%
% \begin{tabular}{p{3.5cm} p{0.75cm} p{1.25cm} p{1.2cm} p{1.2cm} p{1.2cm} p{1.2cm}}
\begin{tabular}{@{}lcccccc@{}}
\toprule
\textbf{Model} & \textbf{S} & \textbf{BN MMLU} & \textbf{BoolQ} & \textbf{CSQA} & \textbf{OBQA} & \textbf{PIQA} \\
\midrule

\multirow{2}{*}{davinci} 
 & 0 & 0.30 & 0.53 & 0.22 & 0.30 & 0.52 \\
 & 5 & - & - & - & - & - \\

\hline

\multirow{2}{*}{Llama-3.2-1b} 
 & 0 & 0.28 & 0.53 & 0.23 & 0.32 & 0.53 \\
 & 5 & 0.28 & 0.58 & 0.23 & 0.32 & 0.54 \\
 
\multirow{2}{*}{Llama-3.2-3b} 
 & 0 & \textbf{0.33} & 0.53 & 0.26 & 0.32 & 0.57 \\
 & 5 & 0.34 & \textbf{0.69} & 0.29 & 0.32 & 0.57 \\

\hline

\multirow{2}{*}{Gemma-2-2b} 
 & 0 & 0.29 & 0.56 & 0.26 & \textbf{0.34} & 0.56 \\
 & 5 & 0.32 & 0.60 & 0.28 & 0.33 & 0.56 \\

 \hline

\multirow{2}{*}{Qwen-2.5-0.5b} 
 & 0 & 0.30 & 0.53 & 0.21 & 0.31 & 0.54 \\
 & 5 & 0.31 & 0.58 & 0.22 & 0.30 & 0.53 \\

\multirow{2}{*}{Qwen-2.5-1.5b} 
 & 0 & \textbf{0.33} & \textbf{0.62} & 0.23 & 0.29 & 0.53 \\
 & 5 & \textbf{0.35} & 0.68 & 0.23 & 0.30 & 0.52 \\

 \hline

\multirow{2}{*}{SmolLM2-135m} 
 & 0 & 0.23 & 0.53 & 0.22 & 0.31 & 0.52 \\
 & 5 & 0.23 & 0.51 & 0.21 & 0.30 & 0.52 \\

\multirow{2}{*}{SmolLM2-360m} 
 & 0 & 0.25 & 0.53 & 0.20 & 0.30 & 0.54 \\
 & 5 & 0.24 & 0.52 & 0.21 & 0.29 & 0.53 \\

\multirow{2}{*}{SmolLM2-1.7b} 
 & 0 & 0.29 & 0.53 & 0.22 & 0.31 & 0.53 \\
 & 5 & 0.30 & 0.55 & 0.21 & 0.30 & 0.53 \\

 \hline

\multirow{2}{*}{BLOOM-560m} 
 & 0 & 0.23 & 0.53 & 0.26 & 0.31 & 0.54 \\
 & 5 & 0.23 & 0.53 & 0.26 & 0.28 & 0.54 \\

\multirow{2}{*}{BLOOM-1b1} 
 & 0 & 0.26 & 0.56 & 0.27 & 0.31 & 0.54 \\
 & 5 & 0.23 & 0.58 & 0.27 & 0.31 & 0.55 \\

\multirow{2}{*}{BLOOM-1b7} 
 & 0 & 0.27 & 0.53 & 0.27 & 0.32 & 0.55 \\
 & 5 & 0.27 & 0.59 & 0.30 & 0.31 & 0.56 \\

\multirow{2}{*}{BLOOM-3b} 
 & 0 & 0.26 & 0.53 & 0.27 & 0.33 & \textbf{0.58} \\
 & 5 & 0.23 & 0.53 & 0.32 & 0.31 & 0.58 \\

 \hline

\multirow{2}{*}{BongLLaMA-3.2-1b} 
 & 0 & 0.25 & 0.53 & 0.22 & 0.33 & 0.52 \\
 & 5 & 0.26 & 0.53 & 0.24 & 0.31 & 0.53 \\

\multirow{2}{*}{BongLLaMA-3.2-3b} 
 & 0 & 0.30 & 0.53 & 0.21 & 0.27 & 0.51 \\
 & 5 & 0.33 & 0.54 & 0.20 & 0.29 & 0.50 \\

 \hline

\multirow{2}{*}{TituLLM-1b-v2.0} 
 & 0 & 0.25 & 0.53 & 0.26 & 0.32 & \textbf{0.58} \\
 & 5 & 0.25 & 0.51 & 0.28 & 0.33 & 0.57 \\

\multirow{2}{*}{TituLLM-3b-v2.0} 
 & 0 & 0.25 & 0.53 & \textbf{0.28} & 0.32 & \textbf{0.58} \\
 & 5 & 0.25 & 0.54 & \textbf{0.33} & \textbf{0.35} & \textbf{0.60} \\

\bottomrule
\end{tabular}
}
\vspace{-0.2cm}
\caption{Benchmark results (normalized accuracy) across models and datasets for 0-shot and 5-shot settings. S: Shots, BN MMLU: Bangla MMLU, davinci: GPT-davinci-002, }
\label{tab:benchmark-results}
\vspace{-0.2cm}
\end{table}

\noindent  \textbf{Bangla MMLU:} In the 0-shot setting, both the TituLLM-1b and TituLLM-3b models score 0.25, placing them in the mid-range relative to other 1b--3b models in the Bangla MMLU benchmark. Neither model shows gains when moving to the 5-shot setting (both remain at 0.25), suggesting that additional examples do not substantially improve performance for this specialized knowledge benchmark. It is possible that the domain-specific knowledge required for MMLU-like tasks is not adequately captured by our model. The primary reason behind this can be the lack of extensive pertaining. We have trained our model with only $\sim37b$
% \textit{36.8b} 
tokens for one epoch. As a result, the model could not capture the full knowledge base \cite{hoffmann2022trainingcomputeoptimallargelanguage}. Another reason behind this can be diversity in datasets. For example, Llama and Qwen models are trained on a high volume of English datasets that have helped these models to have a better knowledge base. 

% Another reason behind this can be seen in the detailed performance analysis (Appendix: ) for the Bangla MMLU dataset. It can be noticed that Qwen2.5-1.5b has done exceptionally well in some topics like English and ICT. If we analyze the questions related to these topics, we will find that many questions contain a mixture of Bangla and English words. As Qwen2.5-1.5b is trained on a high volume of English datasets, the inherent knowledgebase of the model helps it to relate the answer to the question properly. 

\noindent  \textbf{BoolQ:} The BoolQ dataset measures the performance of the model for yes/no question-answering in Bangla. TituLLM-1b achieves 0.53 in the 0-shot setting but drops slightly to 0.51 in the 5-shot setting. In contrast, TituLLM-3B moves from 0.53 (0-shot) to 0.54 (5-shot). However, Llama-3b and Qwen-2.5-1.5b have done much better in this task. As the context length for all BoolQ data matched that of a News document, TituLLM's performance may drop on long contexts. This suggests further pretraining should target longer contexts \cite{chowdhery2022palmscalinglanguagemodeling,kaplan2020scalinglawsneurallanguage}.
% As the context length for all BoolQ data was as large as a News document, TituLLM's performance may drop for long contexts. It suggests further pertaining to the model should be done with long contexts \cite{chowdhery2022palmscalinglanguagemodeling,kaplan2020scalinglawsneurallanguage}.

\noindent  \textbf{CSQA, OBQA, and PIQA:} Commonsense reasoning tasks often challenge smaller-scale language models. The accuracy of the 3B variant of TituLLM, starts at 0.28 (0-shot) and exhibits a more pronounced jump to 0.33 (5-shot) which is the maximum among all models. TituLLM-1b has also shown decent performance on the CSQA dataset. 

OBQA requires both textbook knowledge and reasoning. Similar to CSQA, TituLLM-3b shows superior performance in this dataset 0.35. Both the dataset's results suggest that TituLLM's reasoning capability is better than other base models.

PIQA tests physical commonsense knowledge. TituLLM-3b model shows better performance in this task too with an accuracy of 0.60. By observing the results on the CSQA, OBQA, and PIQA datasets we can say that the model has captured Bangla Language specific reasoning well in spite of being trained with a smaller dataset than others but the results from MMLU and BoolQ shows the impact of limited training. 

\begin{figure}[t]
  \centering
\includegraphics[width=1.0\columnwidth]{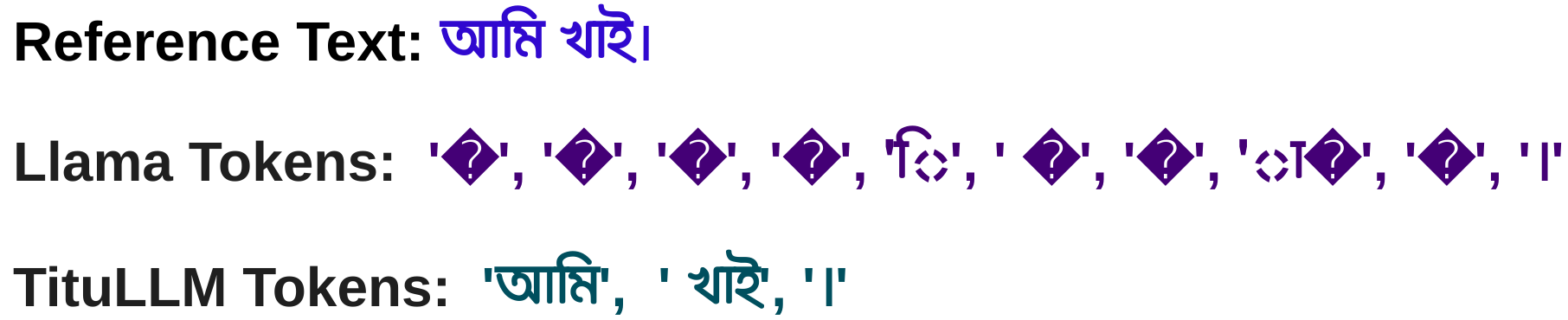}
  \caption{Example of tokenization of Llama and TituLLM tokenizers.}
  \vspace{-0.2cm}
  \label{fig:tokens}
  \vspace{-0.2cm}
\end{figure}

\noindent \textbf{Performance of Tokenizer: } The superior performance of our models in reasoning tasks is mainly an impact of our extended tokenizer. To justify this, we can observe the results of the BongLLaMa models. These models are continual pretrained models with existing open-source Bangla text corpus. If only the dataset could improve the performance then that would be reflected in BongLLaMA models. However, we observe that they are performing similarly to Llama models. To have an interpretation of our extended tokenizer's performance we can look into Figure~\ref{fig:tokens}. The figure shows Llama tokens and TituLLM tokens for a simple sentence in Bangla with two of the most common words. We observe that Llama tokenizer splits the text into character and byte levels. On the other hand, TituLLM tokenizes the sentence into word or subword levels. As a result, TituLLM can deliver more meaningful tokens than Llama for Bangla text. This is an important advantage of TituLLM that not only enables TituLLM to perform better with smaller datasets but also ensures low latency during inference.

% \subsection{Comparison and Trends}

% % \begin{itemize}[noitemsep,topsep=0pt,labelsep=.5em]
% \noindent \textbf{Parameter Scaling Effects.} Across nearly all tasks, the 3B variant outperforms or equals the 1B model. The most pronounced gains appear in tasks requiring deeper commonsense reasoning and knowledge retrieval (e.g., CommonsenseQA BN and OpenBookQA BN).

% \noindent \textbf{Few-Shot Sensitivity.} Although the 1B model remains stable or decreases slightly on some tasks (e.g., BoolQ BN and PIQA BN), the 3B model typically extracts additional benefits from 5-shot prompts. This discrepancy likely stems from higher representational capacity in the larger model.

% \noindent \textbf{Task-Specific Performance.} Neither Titulm model displays strong improvement on Bangla MMLU when given 5-shot examples. For more specialized or domain-specific questions, a small number of examples may be insufficient. Future work could explore larger sets of exemplars or specialized fine-tuning to address domain expertise gaps.
% % \end{itemize}

%% file: sections/related_work.tex
\section{Related Work}
\label{sec:related_work}
% Recent progress in Natural Language Processing (NLP) has highlighted the remarkable evolution of large language models (LLMs), which have transformed computational linguistics capabilities across a diverse array of languages. Nonetheless, the primary focus remains on high-resource languages, often overshadowing advancements in low-resource languages such as Bangla, which is spoken by almost 30 billion of people worldwide. This section explores the latest efforts aimed at improving Bangla language processing by leveraging the power of LLMs. 

% Recent advances in NLP through the development of large language models (LLMs) have shown considerable promise, especially for low-resource languages such as Bangla. This section reviews the literature surrounding the pretraining, instruction tuning, and evaluation of these models, particularly focusing on their application to the Bangla language.

% \subsection{Corpus}

\noindent
\paragraph{Pretraining:}
Pretraining LLMs on Bangla has involved the development of specialized models like BongLLaMA \cite{zehady2024bongllama}, which has been adapted from Llama to better understand and generate Bangla text. The pretraining phase typically leverages large-scale Bangla corpora to improve the model's foundational understanding of the language's syntax and semantics. For instance, \citet{zehady2024bongllama} focused on developing a robust model by pretraining on diverse Bangla data sources, significantly improving the model's performance on native text. Similar efforts have been made in previous research, such as BanglaBERT~\cite{bhattacharjee2022banglabert} and SahajBERT~\cite{diskin2021distributed}, where models underwent extensive pretraining on curated Bangla datasets to better capture linguistic nuances.

% https://arxiv.org/pdf/2406.14670

\noindent
\paragraph{Enhancing Tokenization:}
The evolution of token adaptation in NLP has progressed from linguistic cues and statistical methods~\cite{creutz2006morfessor, luong2013better, zhang2023ask} to phrase-level segmentation~\cite{koehn2007moses, koehn2003statistical}. The rise of deep learning shifted the focus to subword-level segmentation, enhancing the handling of rare words~\cite{sennrich2015neural, kudo2018subword, kudo-richardson-2018-sentencepiece}. More recent efforts emphasize integrating specialized vocabularies into pre-trained LLMs, prioritizing tokenization quality and cost-effectiveness~\cite{ahia2023all, zhang2023ask, zhang2024counting, tejaswi2024exploring}.
\citet{liu2023task} propose a model-agnostic approach for adapting extended vocabularies to LLMs by integrating task-specific vocabularies, prioritizing new tokens, and initializing their embeddings using averaged subword representations. \citet{cui2023efficient} extend Llama's existing vocabulary by incorporating an additional 20k Chinese tokens, enhancing its ability to understand and generate Chinese text. \citet{chiappeoptimizing} develop an adaptive tokenization algorithm that implements a dynamic tokenization dictionary within the Llama model, updating tokens in real-time based on frequency and contextual relevance.

\noindent
\paragraph{Cross-Lingual Model Adaptation:}
Cross-lingual transfer enables models trained in one language to adapt to others without retraining from scratch. Key adaptation techniques include embedding initialization, transliteration, and vocabulary extension.
\citet{jaavid2024romansetu} used transliteration to convert non-Latin languages into Latin scripts for better knowledge transfer. \citet{zhao2024llama} trained a model on millions of target-language tokens without vocabulary extension, achieving performance comparable to models trained on billions of tokens. However, tokenization mismatches reduced inference efficiency.
Studies by \citet{csaki2023efficiently, cui2023efficient, raffel2020exploring, lin2024mala} found that vocabulary extension improves performance while reducing computational inefficiencies. \citet{tejaswi2024exploring} further explored language-specific LLMs, highlighting trade-offs in adaptation for low-resource languages. Their findings emphasize that while vocabulary expansion enhances efficiency, selecting the right base model and vocabulary size is crucial.

\noindent
\paragraph{Benchmarking and Evaluation.}
Evaluating LLMs requires benchmarking datasets that assess a wide range of capabilities and tasks. For Bangla, most existing datasets focus on standard NLP tasks. The BanglaNLG benchmark dataset~\cite{bhattacharjee2023banglanlg} addresses this by integrating six distinct datasets designed to evaluate various aspects of natural language generation (NLG) in Bangla, including Machine Translation, Text Summarization, Question Answering, Multi-turn Dialogue, News Headline Generation, and Cross-lingual Summarization.
Beyond NLG, the Region-Specific Native-QA dataset~\cite{hasan2024nativqa} was developed to assess the question-answering capabilities of leading LLMs, such as GPT-4o, GPT-4, Gemini, Llama-3, and Mistral. By focusing on regionally relevant queries, this dataset ensures that models are tested in real-world Bangla language contexts.
For a broader evaluation of LLMs across multiple tasks, BenLLM~\cite{kabir2024benllm} provides the most comprehensive comparison of model performance. This study benchmarks LLMs against other pretrained models using datasets from diverse sources, offering insights into their strengths and limitations across various NLP tasks. Other relevant benchmarking efforts include sentiment analysis~\cite{hasan-etal-2024-zero}, question answering~\cite{hasan2024nativqa,shafayat-etal-2024-benqa,alam2025nativqaframework}, summarization~\cite{tanjila-etal-2025-bengali}, and cultural understanding~\cite{kabir2025bluck}. 

There remains a notable scarcity of benchmarking datasets for evaluating the emergent capabilities of LLMs, such as world knowledge and cognitive reasoning. To address this gap, we introduce \textbf{\textit{five benchmarking datasets}}, each designed to assess specific competencies, including world knowledge and commonsense reasoning.

%% file: sections/appendix.tex
\section{Pretraining Data}

 \subsection{Tokens in \titu{}}
Table \ref{tab:titulm_tokens} presents the token distribution used for pretraining \titu{}, which are collected from various sources. The largest portion, amounting to 18.80 billion tokens, originates from a deduplicated corpus combining Common Crawl, Books, and Sangraha data. Additionally, synthetic data generated for the model contributes 7.06 billion tokens, comprising translated (1.47b), romanized (3.87b), conversational (0.42b), and audio-transcribed (1.30b) text. The Sangraha dataset further adds 10.94 billion tokens, split between translated (4.26b) and romanized (6.68b) subsets. In total, the \titu{} pretraining corpus consists of $\sim37$ billion tokens, incorporating both natural and synthetic data sources to enhance linguistic diversity and representation.

\begin{table}[h]
    \centering
    \begin{tabular}{p{4cm} c}
        \hline
        \textbf{Data} & \textbf{ \# Tokens (B)} \\
        \hline
        Common Crawl, Books, and Sangraha & 18.80 \\ \midrule
        Synthetic & \\
        \quad Translated & 1.47 \\
        \quad Romanized & 3.87 \\
        \quad Conversation & 0.42 \\
        \quad Audio Transcription & 1.30 \\ \midrule
        Sangraha & \\
        \quad Translated & 4.26 \\
        \quad Romanized & 6.68 \\
        \hline
        \textbf{Total} & \textbf{36.80} \\
        \hline
    \end{tabular}
    \caption{Token distribution for \titu{} pretraining, including contributions from natural and synthetic data sources.}
    \label{tab:titulm_tokens}
\end{table}

\subsection{Rules for Data Filtering}
\label{ssec:rule_data_filtering}
% \todo[inline]{to add}

For pretraining data filtering, we applied a set of carefully designed hand-crafted rules, which are outlined below.

\noindent \textbf{Line Ending with Terminal Punctuation:}  
Determines whether a line concludes with a terminal punctuation mark, including \texttt{``.''}, \texttt{``!''}, \texttt{``?''}, and \texttt{``”''}. Helps assess sentence completeness and filter out incomplete or malformed content.

\noindent \textbf{Line Word Numbers:}  
Calculates the number of words in each line after normalization. Provides insights into whether text consists of single-word lines, short fragments, or full-length sentences.

\noindent \textbf{Line Start with Bullet Points:}  
Identifies lines starting with bullet points, including Unicode symbols like \texttt{\textbackslash u2022}, \texttt{\textbackslash u2023}, \texttt{\textbackslash u25B6}, and others. Useful for recognizing and handling structured lists.

\noindent \textbf{Line Numerical Character Fraction:}  
Measures the proportion of numerical characters in each line. Helps identify lines dominated by numbers, such as statistics, mathematical expressions, or financial reports.

\noindent \textbf{Is Adult URL:}  
Flags documents originating from adult content URLs for filtering inappropriate or explicit content.

\noindent \textbf{Document Languages Identification:} 
Uses FastText \cite{grave2018learning} for language identification, extract language percentages, and filter documents where Bangla meets the specified threshold.

\noindent \textbf{Document Sentence Count:}  
Counts the number of sentences using BNLP Tokenizer \cite{sarker2021bnlp}  for Bangla and NLTK Tokenizer \footnote{\url{}}for English , and it helps to assess document length and complexity.

\noindent \textbf{Document Word Count:}  
Computes the total number of words after normalization. Provides an overall measure of document length and verbosity.

\noindent \textbf{Document Mean Word Length:}  
Calculates the average length of words after normalization. Longer words often indicate more sophisticated vocabulary.

\noindent \textbf{Document Word to Symbol Ratio:}  
Determines the ratio of symbols (\texttt{``\#''}, \texttt{``...''}, \texttt{``…''}) to words. A high ratio may indicate unconventional formatting or non-standard text.

\noindent \textbf{Document Fraction End with Ellipsis:}  
Computes the fraction of lines ending with an ellipsis (\texttt{``...''}, \texttt{``…''}), which may suggest incomplete thoughts or trailing sentences.

\noindent \textbf{Document Unique Word Fraction:}  
Measures the fraction of unique words in a document, providing insight into vocabulary diversity and repetition.

\noindent \textbf{Document Unigram Entropy:}  
Calculates the entropy of the unigram distribution, measuring lexical variety using the formula:
\[
\sum \left(-\frac{x}{\text{total}} \log \frac{x}{\text{total}}\right)
\]
where \( x \) represents counts of unique words in the normalized content. Higher entropy suggests greater lexical diversity.

\noindent \textbf{Document Stop Word Fraction:}  
Determines the ratio of stop words (e.g., ``the,'' ``and,'' ``is'') to total words. A high ratio may indicate informal text, while a low ratio may suggest technical or keyword-dense content.

\noindent \textbf{Fraction of Characters in Top N-Gram:}  
Measures the proportion of characters within frequently occurring word n-grams. Helps assess text repetitiveness and structure.

% \noindent \textbf{Fraction of Characters in Duplicate Word N-Gram:}  
% Calculates the fraction of characters contained within repeated word n-grams, helping to identify redundant or excessively repetitive content.

% \noindent \textbf{Document Top Language Detection:}  
% Determines the primary language of a document, ensuring content is processed in the correct linguistic context and aiding in language-based filtering.

\noindent \textbf{Document Content Classification:}  
Categorizes document content based on profanity, vulgarity, or toxicity to filter inappropriate material.

\noindent \textbf{Document Bad Words Count:}  
Counts offensive or inappropriate words, serving as a stricter filter for explicit content.

% \noindent \textbf{Document Curly Bracket Ratio:}  
% Calculates the ratio of curly brackets (\texttt{``\{''}, \texttt{``\}''}) to total characters. A high ratio may indicate code, structured data, or formulaic content.

\noindent \textbf{Document Bracket Ratio:}  
Determines the ratio of all bracket types (e.g., \texttt{``()''}, \texttt{``[]''}, \texttt{``\{\}''}) to total characters. Useful for detecting technical or structured text such as programming code, mathematical expressions, or legal documents.

 By leveraging language-specific tokenizers and tools, we ensured that the evaluation framework effectively captured the characteristics and patterns unique to Bangla text, enabling robust filtering and alignment with intended use cases.

\subsection{Rules for Cleaning OCR-Extracted Text}
\label{ssec:app_rules_google_ocr}
\begin{figure*}[h]
\vspace{-12cm}
    \centering
    \includegraphics[width=0.6\linewidth]{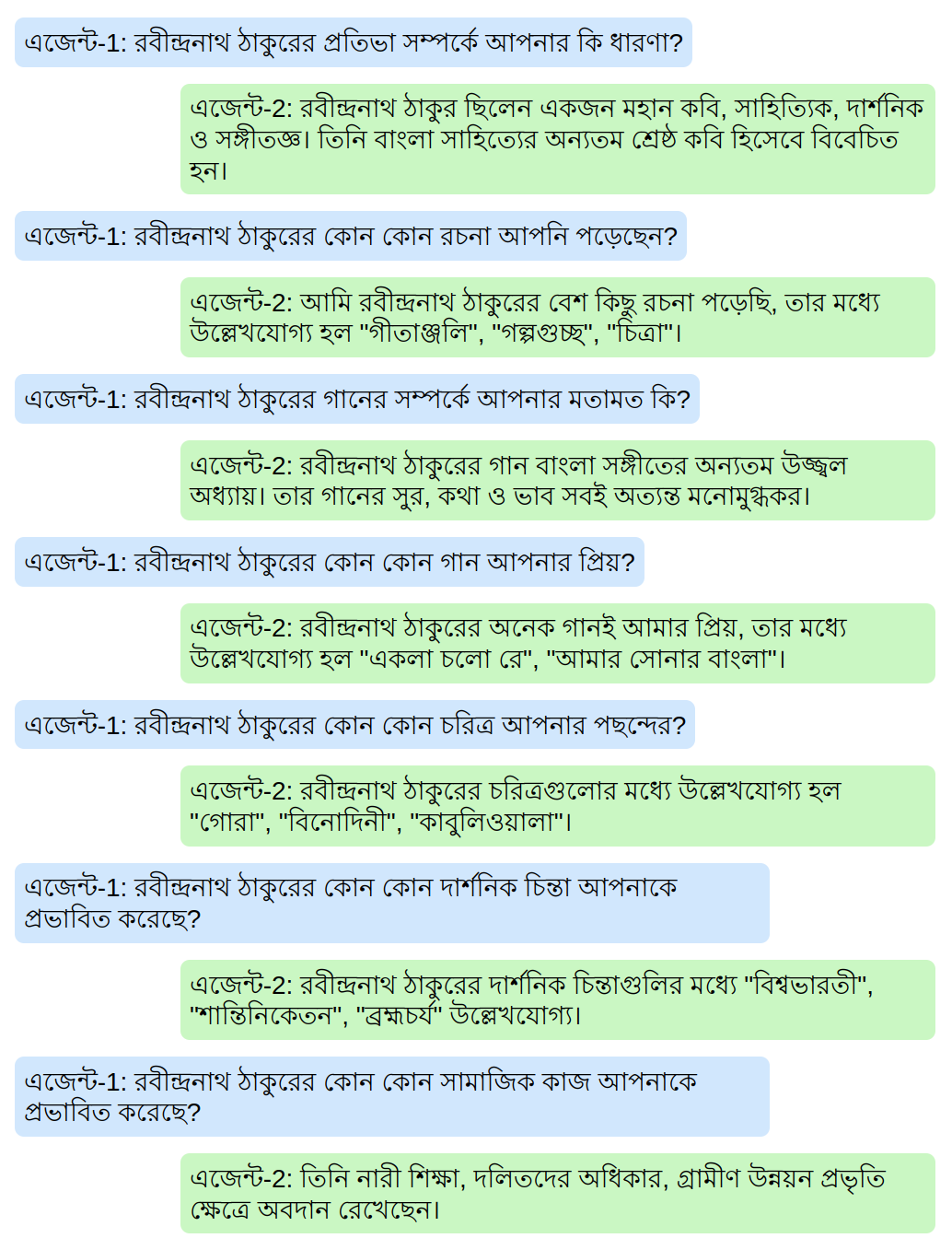}
    \vspace{-0.3cm}
    \caption{Conversation example between two agents on Rabindranath Tagore's contributions to Bangla art.
    }
    \label{fig:data_collection_final}
    % \vspace{-0.3cm}
\end{figure*}

% \todo[inline]{to add}
% We have collected text data from a diverse range of Bangla books, covering genres such as novels, essays, poetry, and academic materials. 
Figure~\ref{fig:data_collection_final} provides an overview of the book data collection process. 
% The following steps outline the procedure for obtaining the final version of the raw text data from these books.
We applied the following rules to filter the text from the OCR-extracted text. 

\begin{itemize}[noitemsep,topsep=0pt,labelsep=.5em] 
\item \textbf{Use of KenLM \cite{heafield2011kenlm}:} KenLM is an efficient statistical language modeling toolkit commonly used for constructing n-gram language models. We trained a language model using high-quality text data, which enabled us to calculate word and sentence scores for the OCR-extracted text. A histogram of these scores was plotted, and thresholds were set based on the distribution to identify poorly recognized sections. The threshold for filtering low-quality text was set at 95\%, meaning that only 95\% of the text was retained.

% The threshold for filtering low-quality text was set at 95\%, meaning that only text with a score above the 95th percentile of the histogram was retained. Mathematically, this can be expressed as:
% \[
% \text{Score Threshold} = \text{Percentile}(95)
% \]
% Where the \(\text{Percentile}(95)\) function returns the score value at the 95th percentile of the word and sentence scores.
    \item \textbf{Word and Sentence Count in Documents:} We calculated the average number of words and sentences per page in the collected books. A minimum threshold for these counts was established to help filter out books with low-quality text. 
    \item \textbf{Percentage of Correct Bangla Words:} We compiled a list of common Bangla words and computed the percentage of these words in each book. A threshold was determined based on the overall distribution of Bangla word occurrences across the corpus.
\end{itemize}

Once these thresholds were established, they were applied across the entire dataset, resulting in the filtering of approximately 50\% of the raw text data.

\section{Tokenizer Details}
Table \ref{tab:tokens-per-word} presents the Tokens per Word (TPW) values for different variants of the Llama-3.2 model. The base Llama-3.2 model has the highest TPW at 7.8397, while the extended Llama-3.2-plus models, with varying context lengths (32K to 96K), exhibit progressively lower TPW values. 

\begin{table}[h!]
\centering
\begin{tabular}{lc}
\hline
\textbf{Model} & \textbf{TPW} \\
\hline
\verb|Llama-3.2|       & 7.8397 \\
\verb|Llama-3.2-plus-32K|     & 2.1346 \\
\verb|Llama-3.2-plus-48K|     & 1.9029 \\
\verb|Llama-3.2-plus-64K|     & 1.7946 \\
\verb|Llama-3.2-plus-80K|     & 1.7370 \\
\verb|Llama-3.2-plus-96K|     & 1.7034 \\
\hline
\end{tabular}
\caption{Tokens per word (TPW) for different Llama models.}
\label{tab:tokens-per-word}
\end{table}

\section{Translation Model (For pertaining data) Evaluation}

\textbf{Scalar Quality Metrics (SQM): } We use Scalar Quality Metrics (SQM) \cite{lankford2022human} to perform human-level evaluation of machine translation outputs in a structured and interpretable manner. SQM employs a 0 to 6 scale to assess the quality of each translated segment, considering both the semantic accuracy and grammatical correctness within the context of the document. This scale provides annotators with a simplified alternative to the traditional 0–100 WMT scoring method, while still capturing nuanced quality differences. Key levels on the scale (0, 2, 4, 6) represent distinct degrees of translation quality—from complete nonsense (0) to perfect translations (6)—with intermediate scores (1, 3, 5) allowing for more precise evaluation. By using SQM, we ensure consistent and context-aware human judgments across different translation models.

% \textbf{Methodology: }

To evaluate translation models, we have collected 100 sentences from the BBC News Dataset \cite{10.1145/1143844.1143892}. We have divided the samples into five categories: Business, Entertainment, Politics, Tech, and Sport. We have 20 samples for each category. After that, we provided the translations from multiple models to two human evaluators (non-authors), and they scored the translations coming from each model.

\label{sec:translation_model_eval}

\section{Expressive Semantic Translation (EST)}
\label{sec:app_est}

\input{sections/EST_lite}

We evaluate EST method on a benchmarking test dataset by both well-known evaluation methods and LLM-based methods. 

\subsection{Evaluation}
\noindent  \textbf{Data Selection: } 
The dataset that has been utilized here for evaluation was taken from \cite{hasan2020not} where a customized sentence segmenter was used for Bangla and two novel methods, namely aligner ensembling and batch filtering, were used to develop a high-quality parallel corpus of Bangla and English with 2.75 million sentence pairings. The 1000 pairs that comprise up the test set of this data were created with extensive quality control and used in this assessment.   

\noindent \textbf{Evaluation metrics: } We employ the BLEU Score \cite{papineni2002bleu}, SacreBLEU Score \cite{post2018call}, BERT Score (F1) \cite{zhang2020bertscore}, and an LLM-Based Evaluation.  

\textbf{LLM-based evaluation: } This approach uses a large language model (GPT-4o) to assess translations qualitatively. The LLM is instructed by the given prompt to assess a Bangla translation against a reference text using the following criteria: accuracy (measures translation accuracy and semantic richness), fluency, readability, and faithfulness. A score between 1 and 10 is then assigned by the LLM, along with a rationale for each translation. Finally, an average score is computed for each translation method. LLM-based evaluation is more flexible and human-like than token-based methods since it may assess more semantic variations and fluency.

\noindent \textbf{Result: } Our comparison of the EST method against industry-standard models like Google Translation API and advanced systems like GPT-4o and Gemini highlights EST's superior performance. As shown in Table \ref{tab:translation_models}, EST leads with remarkable scores in BLEU (\textbf{0.57}), SacreBLEU (\textbf{49.50}), and BERTScore (F1) (\textbf{0.93}). These metrics underscore EST's unparalleled accuracy and contextual richness, with a notable \textbf{20-point} lead in SacreBLEU over the closest competitor. These results affirm the efficacy of EST's iterative refinement in elevating translation quality.

\begin{table}[t]
\centering
\setlength{\tabcolsep}{4pt}
\scalebox{0.85}{
\begin{tabular}{lcccc}
\toprule
\textbf{Translator} & \textbf{BLEU} & \textbf{SBLEU} & \textbf{BS (F1)} & \textbf{LLM Score} \\
\midrule
Google         & 0.33 & 29.00 & 0.91 & \textbf{8.96} \\
GPT-4o         & 0.26 & 22.05 & 0.90 & 8.91 \\
IndicTrans2    & 0.27 & 23.74 & 0.90 & 8.73 \\
Gemini-1.5-pro & 0.31 & 27.08 & 0.89 & 8.80 \\
EST            & \textbf{0.57} & \textbf{49.50} & \textbf{0.93} & 8.95 \\
\bottomrule
\end{tabular}
}
\caption{Comparison of different translation models.}
\label{tab:translation_models}
\end{table}

% First table: Single-column (inside minipage or normal table)

\section{Conversation Data Generation Prompt}
Our methodology for generating Bangla conversational texts involves two specialized roles: the Junior Content Writer and the Senior Content Writer as highlighted in Figure \ref{tab:agent_details}. The Junior initiates dialogues based on culturally significant topics. The Senior meticulously reviews these texts to ensure grammatical precision and enhance quality. This structured approach enables replicable, high-standard conversation generation for NLP research. 

\begin{table}[t]
\vspace{-3.9cm}
\centering
\setlength{\tabcolsep}{3pt} 
\scalebox{0.78}{%
\begin{tabular}{p{2cm}|p{2.5cm}|p{4.5cm}}
\toprule
\textbf{Component} & \textbf{Description} & \textbf{Details} \\ \midrule
Input Text Snippet & Bangla text snippet provided to initiate the conversation. & The text is a prompt that provides a topic for conversation, such as ``Rabindranath Tagore's contributions to Bengali art'', sourced from educational and cultural databases like Wikipedia or Banglapedia. This serves as the basis for the conversation generation task. \\ \midrule
Junior Content Writer & Agent tasked with generating the initial conversation. & \textbf{Role:} Content Creator \newline \textbf{Goal:} To generate high-quality, coherent, and engaging conversational text in Bangla. \newline \textbf{Backstory:} As an expert in Bangla language conversation generation, your responsibility includes initiating and maintaining a dialogue that is both interesting and relevant to the given topic. \newline \textbf{Capabilities:} Produces verbose outputs to ensure detailed and extensive dialogue. No delegation is allowed. Employs a language learning model (LLM) without additional tools. \\ \midrule
Senior Content Writer & Agent tasked with reviewing and refining the conversation. & \textbf{Role:} Examiner \newline \textbf{Goal:} To edit and enhance the quality of Bangla conversational text, ensuring it meets high standards of grammar and coherence. \newline \textbf{Backstory:} You review the conversation generated by the Junior Content Writer, focusing on grammatical accuracy, linguistic quality, and content relevance, making necessary revisions to uphold quality standards. \newline \textbf{Capabilities:} Provides verbose feedback and detailed edits, uses a language model (LLM) for text correction and improvement, and operates independently without delegation. \\ \bottomrule
\end{tabular}
}
\caption{Detailed roles and tasks of agents for generating Bangla conversational text.}
\label{tab:agent_details}
\end{table}

%% file: sections/EST_lite.tex
\subsection{Expressive Semantic Translation}

The EST method innovatively enhances neural machine translation by infusing expressiveness and contextual relevance through an iterative refinement process utilizing LLMs. The EST method encompasses several pivotal steps:

\noindent \textbf{Initial Translation:} A standard translation model \(M\) converts text from a source language \(L_1\) into a preliminary translation \(t_0\) in the target language \(L_2\), which often lacks expressiveness and contextual depth.

\noindent \textbf{Enhanced Linguistic Refinement:} Employing multiple LLMs, the initial translation \(t_0\) is refined into diverse candidate translations that exhibit greater naturalness and idiomatic correctness in \(L_2\).

\noindent \textbf{Quality Diversity:} This phase synthesizes the best elements from the candidate translations through a prompt-based evaluation method, aiming to construct a translation that faithfully represents the nuances of \(L_2\).

\noindent \textbf{Candidate Ranking:} Through model-based evaluations, candidates are assessed for contextual appropriateness and linguistic coherence, with the highest-scoring translation selected as the most suitable.

\noindent \textbf{Final Selection:} After multiple iterations, the optimal translation is selected from the top candidates based on rigorous evaluations, ensuring adherence to the linguistic and contextual standards of \(L_2\).

These steps ensure that the final output not only meets the literal translation requirements but also enhances the translation quality by embodying the naturalness and contextual relevance, significantly surpassing traditional methods.